\documentclass[conference]{IEEEtran}
\IEEEoverridecommandlockouts
\usepackage{cite}
\usepackage{amsmath,amssymb,amsfonts}
\usepackage{algorithm}
\usepackage{algorithmic}
\usepackage{graphicx}
\usepackage{textcomp}
\usepackage{xcolor}
\usepackage{subfigure}
\usepackage{booktabs}
\def\BibTeX{{\rm B\kern-.05em{\sc i\kern-.025em b}\kern-.08em
    T\kern-.1667em\lower.7ex\hbox{E}\kern-.125emX}}
\begin{document}

\title{Learning to Drive via Apprenticeship Learning and Deep Reinforcement Learning
}

\author{\IEEEauthorblockN{1\textsuperscript{st} Wenhui Huang}
\IEEEauthorblockA{\textit{Industrial and Information Engineering} \\
\textit{Politecnico Di Milano}\\
Milano, Italy \\
wenhui.huang@mail.polimi.it}
\and
\IEEEauthorblockN{2\textsuperscript{nd} Francesco Braghin}
\IEEEauthorblockA{\textit{Industrial and Information Engineering} \\
\textit{Politecnico Di Milano}\\
Milano, Italy \\
francesco.braghin@polimi.it}
\and
\IEEEauthorblockN{3\textsuperscript{rd} Zhuo Wang}
\IEEEauthorblockA{\textit{School of communication engineering} \\
\textit{Xidian University}\\
XiAn, China \\
zwang\_ll@stu.xidian.edu.cn}
}

\maketitle

\begin{abstract}
With the implementation of reinforcement learning (RL) algorithms, current state-of-art autonomous vehicle technology have the potential to get closer to full automation. However, most of the applications have been limited to game domains or discrete action space which are far from the real world driving. Moreover, it is very tough to tune the parameters of reward mechanism since the driving styles vary a lot among the different users. For instance, an aggressive driver may prefer driving with high acceleration whereas some conservative drivers prefer a safer driving style. Therefore, we propose an apprenticeship learning in combination with deep reinforcement learning approach that allows the agent to learn the driving and stopping behaviors with continuous actions. We use gradient inverse reinforcement learning (GIRL) algorithm to recover the unknown reward function and employ REINFORCE as well as Deep Deterministic Policy Gradient algorithm (DDPG) to learn the optimal policy. The performance of our method is evaluated in simulation-based scenario and the results demonstrate that the agent performs human like driving and even better in some aspects after training.
\end{abstract}


\section{Introduction}
Recent studies indicate that the interest in applying robotics and autonomous system to real life is growing dramatically \cite{abbeel2010autonomous,silver2016mastering}. Especially, the pace of techinical upgrading and innovation for autonomous vehicle driving is accelerating a lot \cite{silberg2012self} and this is mostly thanks to the capability of the machine learning(ML). 

Reinforcement learning (RL), as one branch of the ML, is the most widely used technique in sequential decision making problem. RL can learn the optimal policy through a process by interacting with unknown environment. RL algorithms have been successfully applied to the autonomous driving in recent years \cite{riedmiller2007learning,sharifzadeh2016learning}. However, these applications are not only limited to the discrete aciton problems but also suffer from "curse of dimensionality" once the action extends to continuous state space. In order to solve large continuous state space problem, deep learning (DL) has been implemented in RL, yielding deep reinforcement learning (DRL) \cite{lillicrap2015continuous}. In recent study, the Deep Deterministic Policy Gradient (DDPG) algorithm which belongs to DRL family, has been successfully applied to target following control \cite{li2017learning}.

One of the issues in RL is the reward function. Knowing that autonomous vehicle driving is not a trivial problem, the reward function is tough to be hand-made directly. To overcome this problem, \cite{abbeel2004apprenticeship} proposed apprenticeship learning via inverse reinforcement learning (IRL) approach . IRL aims at recovering the unknown reward function by observing expert demonstration. 

Both forward driving and stopping under the traffic rules are frequent behaviors in real life driving. However, recent studies \cite{xiong2016combining,yi2018deep} are only focusing on obstacle avoidance and there is no research on learning forward driving and stopping behaviors by considering traffic rules via reinforcement learning techniques. In this paper, we addressed above problem by means of apprenticeship learning in combination with DRL approach. More specifically, we implemented gradient inverse reinforcement learning (GIRL) algorithm \cite{pirotta2016inverse} to recover the unknown reward funciton and employed DDPG algorithm in order to train the agent to drive by keeping traffic rules and stop in front of the stop sign autonomously. Furthermore, REINFORCE algorithm is employed in RL step as well in order to compare the performance with DDPG algorithm.

\section{Related Works}

At the early state, researchers tried to exploit Aritifical Neural Networks as the controller of the taking action. One of the typical paper is ALVINN \cite{pomerleau1989alvinn}. The paper proposed a 3-layers back-propagation network to complete the task of road following. The network takes images from camera as the inputs, passing through 29 hidden layers, and produces the direction of the vehicle should travel along the road as the output. After certain episodes of the training, the car could navigate successfully along the road .

One of the more advanced appliations is utilizing DL technique-convolutional neural networks (CNNs) with Dave2-system which was exactly implemented in \cite{bojarski2016end}. Dave2-system is an end-to-end system which is inspired from ALVINN. The CNN network in this paper consists of 9 layers, including a normalization layer, 5 convolutional layers and 3 fully connected layers . A recent paper \cite{tian2018deeptest} employed CNNs to the motion planning layer as well. 

Although utilizing the DL algorithm directly as the controller of the behavior seems good enough to achieve the target, it belongs to the behavior cloning which means it only has knowledge about the observed data. 
This kind of approach can only be acceptable under a good hypothesis as well as good data including all of the possible cases.

To avoid falling into behavior cloning class, the vehicle should explore and exploit behavior by itself in unknown environment and the approach that is able to handle this case is Reinforcement Learning. Reinforcement Learning is learning what to do-how to map situations to actions-so as to maximize a numeral reward \cite{sutton2018reinforcement}. In  \cite{yu2016deep} a deep Q-network (DQN) algorithm is utilized as a decision maker. By passing through the network with 84 by 84 images, three discrete actions, faster and faster-left as well as faster-right, are returned by the frame. Different from previous paper, \cite{gomez2012optimal} employed dynamic model rather than kinematic with same DQN algorithm. However, both of the applications are still limited to the discrete action space. 

Being aware that driving in real life could not be achieved with several discrete actions, researchers turn to develope continuous control algorithms. One of the popular algorithms that can handle continuous action space problem is Actor-Critic(AC). A paper \cite{schulman2015high} from Berkeley university evaluated AC algorithm on the classic cart-pole balancing problem, as well as 3D robots by tuning with bias and variance. Considering the complexity of sampling and diverse problem of AC, Google Deepmind team published a new upgraded AC algorithm-DDPG \cite{lillicrap2015continuous} in 2016. The paper indicates that DDPG can learn competitive policies using low dimentional observations with only a straightforward actor-critic architecture.

In RL, the reward function plays a significant role since the agent is aiming at getting higher reward whenever it achieves the goal. A classic paper \cite{abbeel2004apprenticeship} published by Standford university proposed IRL algorithm to recover the unknown reward function based on expert's demonstration. Apprenticeship learning has been successfully applied to autonomous vehicles such as learning to drive by maximum entropy IRL \cite{kuderer2015learning} and projection-based IRL \cite{sharifzadeh2016learning}. The bottleneck of the above mentioned methods is the requirement of solving multiple forward RL problems iteratively. A new IRL algorithm stated in \cite{pirotta2016inverse} is gradient inverse reinforcement leanring (GIRL). The idea is to find the reward function that minimizes the gradient of a parameterized representation of the expert's policy based on assumption of reward function is in linear combination with reward features. 

In this paper, we recover the reward function by means of GIRL algorithm and implement DDPG algorithm to learn the optimal policy based on the recovered reward function. REINFORCE algorithm is employed in RL part as well to compare the performance with DDPG algirithm. Moerover, in order to perform human-in-the-loop (HITL), we utilize IPG CarMaker software which is able to interact with driving simulator. Both of the dynamical model of the agent and virtual environment are built in CarMaker and no other road-users are involved in order to fully focus on the driving and stop performance. The experimental results indicate our approach is able to let the agent learn to drive autonomously over continuous actions and the performance is even better than the expert in some aspects.

\section{Preliminaries}
\subsection{Background}
A Markov decision process (MDP) is defined by a tuple, denoted as $\mathcal{M} = \{ \mathcal{S}, \mathcal{A}, \mathcal{P}, \mathcal{R}, \gamma \} $, where $\mathcal{S}$ is state space; $\mathcal{A}$ is action space; $\mathcal{P}$ is transition probability. It stands for the probability of the transition from state $s$ to $s^{'}$ upon taking action $a\in \mathcal{A}$; $\mathcal{R}: \mathcal{S}\to\mathcal{A}$ is the reward (function), it indicates how good the action $a\in \mathcal{A}$ executed from state $s\in \mathcal{S}$ is; And $\gamma$ is discount factor which is limited in the range of [0,1). The policy $\pi$ characterizes the agent's action in MDP problem. More formally, the policy is a mapping from given states to probabilities of selecting each possible action:$\pi(a\mid s) = \mathcal{P}(a=A_{t}\mid s=S_{t})$. The expected retrun based on the state s following the policy $\pi$ is defined as Value funciton, also called state value function, denoted as $V_{\pi}(s)$. In RL, we formalize it in mathematical way:

\begin{equation}
V_{\pi}(s) = \mathbf{E}_{\pi}[R_{t}\mid S_{t}=s]=\mathbf{E}_{\pi}\big[\sum_{k=0}^{\infty}\gamma^{k}r_{t+k+1}\mid S_{t}=s\big]
\end{equation}

Note that in case of terminating state, the value will be 0 always. Similarly, the expected return taking action $\emph{a}$ at state s following policy $\pi$ is defined as Q function, denoted $Q_{\pi}(s,a)$. The Q funciton can be formalized as:

\begin{equation}
\begin{aligned}
Q_{\pi}(s,a) 
& =\mathbf{E}_{\pi}\big[\sum_{k=0}^{\infty}\gamma^{k}r_{t+k+1}\mid S_{t}=s,A_{t}=a\big] 
\end{aligned}
\end{equation}

Furthermore, Many approaches in reinforcement learning make use of the recursive relationship known as the Bellman equation:
\begin{equation}
Q_{\pi}= \mathbf{E}\big[r_{t+1}+\gamma Q_{\pi}(S_{t+1},A_{t+1})\mid S_{t}=s,A_{t}=a\big]
\end{equation}

\subsection{Gradient Inverse Reinforcement Learning}

The logic behind the GIRL algorithm is to find out the reward function by minimizing the gradient of a parameterized representation of the expert's policy. In particular, when the reward function can be represented by linear combination with the reward features, the minimization can be solved effectively with optimization method. Under the assumption of linear combination, it is possible to formalize the reward in the following way:

\begin{equation}
r_{\omega}(s,a)=\varphi^{T}(s)\omega\label{2.63}
\end{equation}

where $\varphi(s)$, $\omega$ $\in \mathcal{R}^{q}$ and q is the dimenstion of the reward features. Considering the expert has his own policy and reward mechanism(still unknown), the objective function could be formalized as :

\begin{equation}
J(\theta^{E},\omega^{E}) =\int_{s}\mathcal{P}(s^{'}\mid s,a)\int_{A}\pi_{\theta}^{E}(s,a)\varphi(s)^{T}\omega^{E}\ dsda\label{2.65}
\end{equation}

where the superscript E represents expert. Since the target of GIRL algorithm is that recovering the reward function as close as the expert's while the expert's policy is completely known, the problem can be formalized as minimizing the $\ell^{2}$-norm gradient of objective function :

\begin{equation}
\omega =\mathop{arg\ min}\limits_{\omega \in \mathbf{R}^{q}} \ \lVert \nabla_{\theta}J(\theta^{E},\omega)\rVert^{2}
\end{equation}

\subsection{Deep Deterministic Policy Gradient}

DDPG algorithm  \cite{lillicrap2015continuous} combines the advantages of the Actor-Critic and DQN \cite{mnih2015human} algorithm so that the converge becomes easier. In other words, DDPG introduces some concepts from DQN, which are employing target network and estimate nework for both of the Actor and Critic. Moreover, the policy of DDPG algorithm is no longer stochastic but deterministic. It means the only real action is outputed from actor network instead of telling probability of different actions. The critic network updating based on the function:

\begin{equation}
\begin{aligned}
L & =\frac{1}{N}\sum_{i}^{N}\big(Q(s_{t},a_{t}\mid \theta^{Q})-y_{i}\big)^{2},\\
\end{aligned}
\end{equation}	

where $y_{i} =r_{i}+\gamma Q^{'}(s_{t+1},a_{t}\mid \theta^{Q^{'}})$ is the Q value estimated by target network and  and N indicates the total number of minibatch size. The actor network is updated by means of gradient term:

\begin{equation}
\nabla_{\theta^{\mu}}J \approx \frac{1}{N}\sum_{i}^{N}\nabla_{a}Q(s,a\mid \theta^{Q})\mid_{s=s_{i},a=\mu(s_{i})}\nabla_{\theta^{\mu}}\mu(s\mid \theta^{\mu})\mid_{s_{i}}\label{2.48}
\end{equation}

Where $Q(s,a \mid \theta^{Q})$ is from critic estimate network. Furthermore, DDPG algorithm solves continuous action space problem by means of two key techniques, namely ”Experience Replay” and ”Asynchronous Updating”.

\section{Our appoach}

In order to implement GIRL algorithm, we performed HITL at the first step. Several policy features are built afterward with extracted states during the HITL and the quality of the designed policy features are checked by means of maximum likelihood estimation (MLE) method. Then, we designed reward features in the sense of desired targets and recovered the weight of the each feature through GIRL algorithm. Having the recovered reward function, we were able to train the agent with REINFORCE and DDPG algorithms at the final step. 

\subsection{Human In The Loop}

\begin{figure}[t]
	\centering
	\includegraphics[width=8cm]{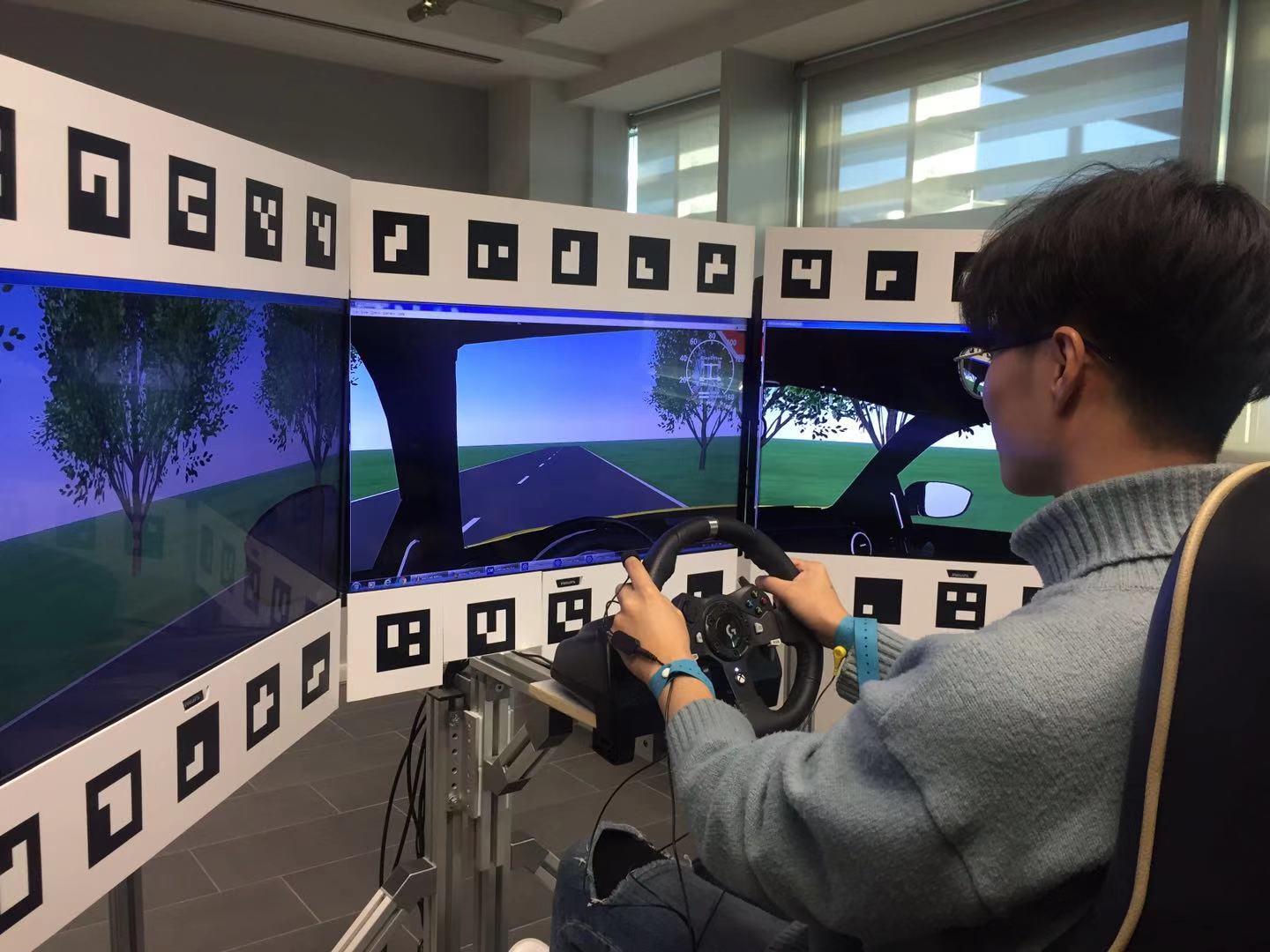}
	\caption{Human In The Loop}
	\label{fig1.0}
\end{figure}

To complete HITL, the expert interfaces with simulator and CarMaker through controlling pedal, braking and steering(Fig. \ref{fig1.0}). The pedal and braking are both limited in the range of [0,1], denoted to $a_{p}\in [0,1]$, $a_{b}\in [0,1]$ respectively. 1 denotes the pedal or braking has been pushed to the maximum while 0 denotes that pedal and braking are totally released. Considering that no one push both of the pedal and braking at the same time in real life, these two actions could be merged as one, denoted as $a_{p}\in [-1,1]$, where [-1,0] means braking and [0,1] means acceleration. Moreover, the steering is limited in the range of $a_{s}\in[-\frac{5}{2}\pi,\frac{5}{2}\pi]$ since the steering wheel in the simulator can rotate 2 and half circle in the maximum. Hence, we can write down these actions as a vector:

\begin{equation}
a \ = [a_{p},a_{s}]^{T}
\end{equation}

Notice that if all of the data are perfect, the vehicle doesn't have perception about penalization since the reward features will be always assigned as 0 (no penalization). Hence we provide 30 trajectories with bad performance among 150 over all  trajectories and consequently a total of 44145 labeled dataset are gathered in the end.

\begin{figure*}[!tp]
	\centering
	\subfigure[Feature 1]
	{\includegraphics[width=5.1cm]{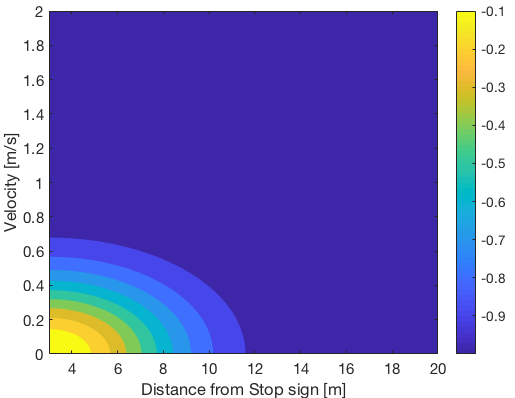}\label{2.0a}}
	\subfigure[Feature 2]
	{\includegraphics[width=5.5cm]{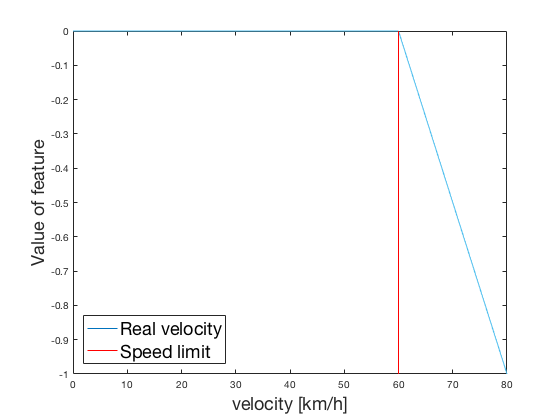}\label{2.0b}}
	\subfigure[Feature 3]
	{\includegraphics[width=5.5cm]{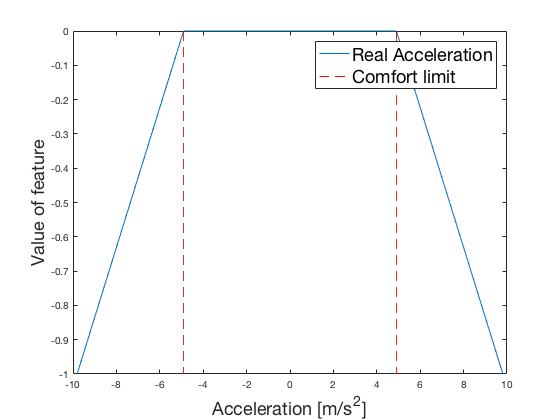}\label{2.0c}}
	\caption{Reward Features}
	\label{fig2.0}
\end{figure*}

\subsection{Policy Features Building}

Inpired from \cite{sutton2018reinforcement}, we assume that the action is in linear combination with policy features $a=\theta^{T}*\phi(s)$, where $\theta$ are the policy parameters and $\phi(s)$ are policy features. The policy features can be states directly from the sensors or constructed by the states. Using the states detected from sensors directly as the policy features may be one kind of solution but in order to shape the action in a smooth way, we selected to build policy features based on the gathered states. The policy features should be built in a sensible way so that they are able to tell the meaningful action w.r.t, the goals. For instance, there should be some features take high values when the vehicle need to accelerate or decelerate hard while some low values in the opposite situation. The overall logic behind designing the policy features are as following:

\begin{enumerate}
	\item Collecting data.
	\item Building the policy features $\phi(s)$ based on the gathered data.
	\item Compute the policy parameters  $\theta_{expert}$ by implementing $\mathit{MLE}$ method
	\item Input the deterministic action $a=\theta_{expert}^{T}*\phi(s)$ to the simulator(CarMaker)
	\item If the vehicle has perception of the target, the features are "good" enough.(e.g. at least the vehicle should perform braking when it is close to stop sign even though the quality of performance may be poor)
	Otherwise, the features are judged as bad. In this case, go back to step 2 and repeat.
\end{enumerate}

By following above logic, 9 features are built at the end and fed to the RL algorithms as the input.

\subsection{Reward Building}

In this study, the reward function is built in the same way as \cite{abbeel2004apprenticeship} proposed. We assume there exists some "true" reward function $R(s)=\omega^{T}\varphi(s)$, where $\omega\in \mathbb{R}^{k}$ and $\vert\vert\omega\vert\vert_{1}\leq 1$ in order to bound the reward in the range of [-1,0]. Since it is linear relationship, the reward features should include all of the aspects w.r.t. following targets:

\begin{enumerate}
	\item The vehicle should stop in front of the stop sign with reasonable distance, not in the middle of the road, not crossing over.
	\item The velocity of the vehicle should not exceed the speed limit, or if it is already higher than the limit, the vehicle should brake at once.
	\item The vehicle should not perform sudden acceleration or emergency braking.
\end{enumerate} 

Therefore, three reward features have been built by following above logics:

$\boldsymbol{\varphi_{1}(s)}$: This feature is built in order to satisfy the demand of stopping in front of the stop sign. There are two indices can be employed to evaluate the performance of the vehicle. First one is vehicle velocity and the other one is distance from the stop sign. A behavior is judged to be poor if the vehicle get null velocity but far from the stop sign or the speed is not zero even if it has reached to the stop sign. To consider both of the factors, we employed multivariate Gaussian distribution function as the first reward feature (Fig. \ref{2.0a}). The mean $\mu$ is a vector with two components that indicates the ideal value of the  velocity and distance from the stop sign, denoted as $\mu=[v_{x}^{*},d_{stop}^{*}]^{T}$.

\begin{equation}
\varphi_{1}(v_{x},d_{stop})=exp\bigg(-\frac{\big(x(s)-\mu)^{T}(x(s)-\mu\big)}{2\sigma^{2}}\bigg)-1
\end{equation}
$\boldsymbol{\varphi_{2}(s)}$: This feature is related to speed limit which is also very important during the driving(Fig. \ref{2.0b}). The vehicle should be punished when it exceeds the allowed speed. To let the vehicle have a better perception, a smooth penalization has been built as:

\begin{equation}
\varphi_{2}(v_{x},v_{lim})=min(0,v_{lim}-v_{x})
\end{equation}

$\boldsymbol{\varphi_{3}(s)}$: Last feature is related to the comfort limit of the vehicle(Fig. \ref{2.0c}). The vehicle should avoid emergency braking not only for the safety but also from the comfort point of view since no other road-users are interfaced with environment. Also in this case, the vehicle is penalized in smooth way with linear relationship:

\begin{equation}
\varphi_{3}(g,acc_{x})=min(0,0.5g-|acc_{x}|)
\end{equation}

By implementing GIRL algorithm with above features, the final recovered weights are:
\begin{equation}
\omega = [0.5512,0.1562,0.2926]^{T}
\end{equation}

\subsection{Reinforcement Learning}

\begin{figure*}[!tp]
	\centering
	\subfigure[Gradient of REINFORCE]
	{\includegraphics[width=4.1cm]{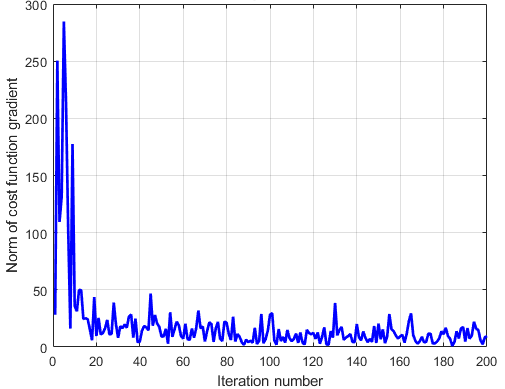}
		\label{3.0a}}
	\subfigure[Reward of REINFORCE]
	{\includegraphics[width=4.1cm]{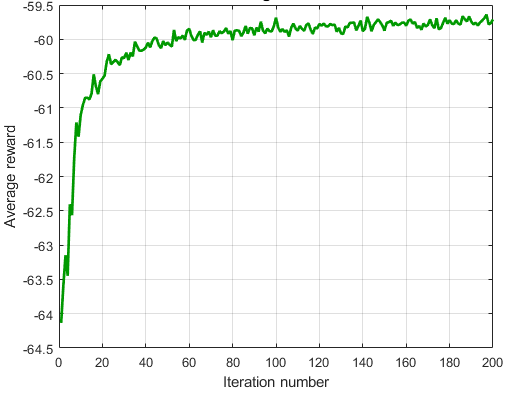}
		\label{3.0b}}
	\subfigure[Critic Loss of DDPG]
	{\includegraphics[width=4.3cm]{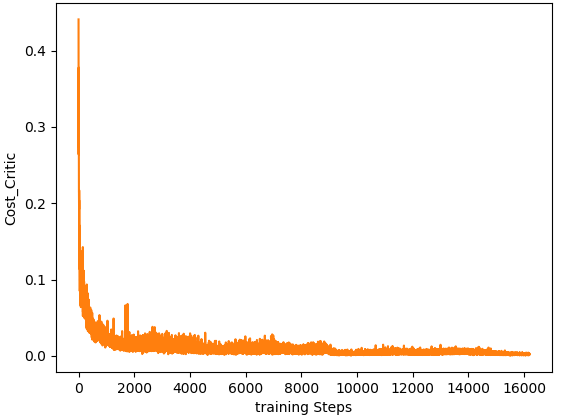}
		\label{3.0c}}
	\subfigure[Reward of DDPG]
	{\includegraphics[width=4.3cm]{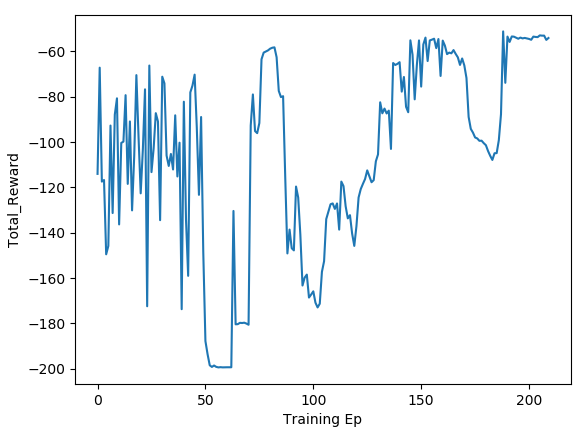}
		\label{3.0d}}
	\caption{Converge of REINFORCE and DDPG}
	\label{fig3.0}
\end{figure*}

\begin{table}
	\centering
	\caption{Hyper-parameters}
	\begin{tabular}{lrr}  
		\toprule
		Hyper-parameters  & REINFORCE & DDPG \\
		\midrule
		Initial Policy Parameter		& $\theta_{expert}$  & $\theta_{random}$      \\
		Discount Factor      & 0.995 & 0.990      \\
		Initial Learning Rate(Actor)    & 0.001  & 0.001      \\
		Initial Learning Rate(Critic)   & -  & 0.0003      \\
		\bottomrule
	\end{tabular}
	\label{table1}
\end{table}

To implement RL algorithms, several hyper-parameters should be defined in the first place. The hyper-parameters utilized in this study can be found in Table \ref{table1}. 
The significant difference between two algorithms is the initial policy parameter.
For REINFORCE, the initial policy parameter is the one recovered from the MLE method while it is randomly initialized for DDPG algorithm. In other words, the agent trained by REINFORCE algorithm has the pre-knowledge about the driving whereas DDPG has to learn from the beginning.

Moreover, one of the most challenging part of the RL is the trade off between exploitation and exploration. If the agent never explores new actions, the algorithm will comverge into poor local minima or even could fail to converge. 
In this study, the exploration is implemented as the Gaussian noise form directly to the action and starts to discount when the counter is larger than the memory size. More specifically, the Gaussian variance starts from 3 and decays to 0 after around 50 episodes.

\section{Experiments}

\subsection{Agent}

In this study, we propose to employ a dynamic rather than kinematic vehicle model in order to let the simulation be more real. Therefore, a classic Volkswagen Beetle model with electrical powertrain is selected from IPG CarMaker software. The rigid body mass is 1498kg and the car equips with four same types of tyres(RT\_195\_65R15). The agent is allowed to perform continuous actions w.r.t. pedal and braking in the range of [0,1]. 0 represents release the pedal/brake totally whereas 1 means maximum push of both actions. Furthermore, multiple sensors like traffic detection sensors, lane detection sensors and so on, are set on the vehicle body in order to gather the information from the environment. 

\subsection{Environment}

Since this study aims at learning forward driving and stopping behavior by keeping several traffic rules, the road is straight forward without any curves. Two traffic signs, speed limit sign and stop sign respectively, are set on the road and the road condtion is regard as normal, which means friction coefficient is equal to 1.0. 

\subsection{Training Strategy}

RL is definitely different from the Behavior Cloning (BC). BC approach recovers the expert optimal policy by learning  the state-action mapping in a supervised way \cite{metelli2017compatible}. In other words, the  policy can be recovered by minimizing the performance difference between the agent and expert. Though this kind of appoach could learn the target in a fast pace, it doesn't hold generalization. More specifically, the policy learnt by BC method will perform poorly once suffers from the states never visited during the training. Therefore, it needs hundreds of data to be fed so that cover all of the possible cases when the environment is stochastic \cite{ho2016generative}. In contrast, given a reasonable reward mechanism, the policy learnt by RL is able to perform well with the states never observed during the training. And it is the exact logic implemened in this study. We fixed the initial velocity of the agent as 60km/h during the training which is the critical value of the speed limit sign. After learning, we checked out the performance of the agent by implementing randomly intialized start velocity and different road length which are never seen before. The empirical results showed that the agent learnt by our approach did achieve the targets with outstanding performance.

\begin{algorithm}[tb]
	\caption{Exploration}
	\label{algorithm1}
	\textbf{Input}: Variance\\
	\textbf{Parameter}: Discount Factor\\
	\textbf{Output}: Discounted Variance
	\begin{algorithmic}[1] 
		\STATE Variance=3
		\STATE Discount Factor=0.999
		\STATE Counter=1
		\FOR{$i \in [0,Max\_epsisode]$}
		\FOR{$j \in [0,Max\_steps]$}
		\STATE $a \gets np.random.normal(a,var)$
		\STATE $Counter \gets Counter + 1$
		\IF {$Counter>Memory$}
		\STATE $Variance \gets Variance \times Discount \ Factor$
		\ENDIF
		\ENDFOR
		\ENDFOR
	\end{algorithmic}
\end{algorithm}

\subsection{Results}

In this section, we provide and analyse the training results of two different RL algorithms.

Fig. \ref{fig3.0} shows the overall converge performance during the training. As one can see from Fig. \ref{3.0a} and \ref{3.0b}, the reward asymptotically converge to a stable value when the gradient of REINFORCE algorithm close to 0. Similary, the reward of DDPG algorithm tends to be stable around the same value as REINFORCE at the end of the iterations with the reduction of the Critic network loss. Specifically, the agent trained by DDPG algorithm used first 50 episodes to fill full the memory and explored new actions with Gaussian noise for further 50 episodes. Therefore, the reward in Fig. \ref{3.0d} bounces up and down from 50th episode to 100th episode. However, the agent did understand how to drive after the noise dacaying to 0 (after 100th episode) and tried to get closer to the stop sign as much as possible. The reduction of the reward from around 160th episode is because the agent got to the stop sign without null velocity. In other words, the agent was trying to figure out what would happen in case of crossing over the stop sign. Qualitatively, the performance of agent is very outstanding after around 190 iterations. Comparing with DDPG, the reason for stable increasement of reward in REINFORCE algorithm is that the initial policy parameter is assigned as $\theta_{expert}$ rather than randomly initialized number. Therefore, the agent already had the pre-knowledge about drving before the training. However, though both of the algorithms converged around 200 iterations but actually the computational cost of REINFORCE is much higher than DDPG. This is because REINFORCE is an off-line updating algorithm which means the sampling efficiency is very poor. Therefore, each of the iteration in REINFORCE process contains 50 trajectories. On the contrary, the single iteration in DDPG algorithm includes only one trajectory thanks to the on-line updating mechanism. Thus, comparing with REINFORCE, DDPG algorithm holds lower computational cost and converges much faster even though it was learning from the beginning. 

\begin{figure}[!tp]
	\centering
	\subfigure[Distance VS Velocity]
	{\includegraphics[width=4.2cm]{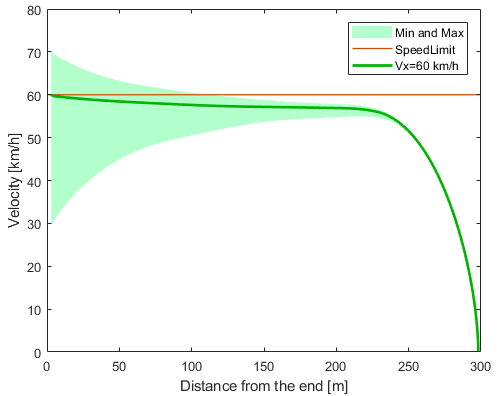}
		\label{4.0a}}
	\subfigure[Distance VS Acceleration]
	{\includegraphics[width=4.1cm]{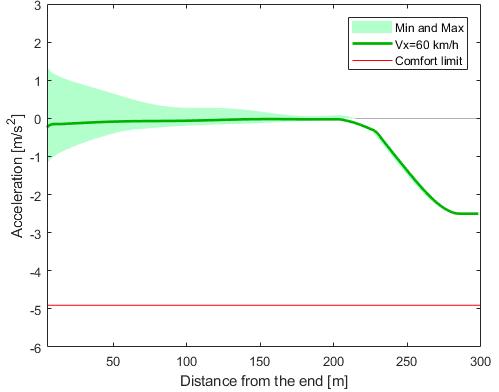}
		\label{4.0b}}
	\caption{Performance of REINFORCE. The shaded area, dark green and red curve denotes the states visited by agent, reference trajectory and critical value for the penalization.}
	\label{fig4.0}
\end{figure}

After training, we checked the performance of the agent by applying different initial velocity and road length. Fig. \ref{fig4.0} demonstrates the overall results of the agent trained by REINFORCE with the start velocity in the range of [30,70]. As seen in the Fig. \ref{4.0a}, the agent is able to maintain the velocity according to the speed limit of the road especially when the initial velocity is already higher than the critical value. Moreover, it did stop in front of the stop sign without performing any emergency braking(Fig. \ref{4.0b}). The overall performance of this agent is very similar as the expert's during the HITL.

\begin{figure}[!tp]
	\centering
	\subfigure[Distance VS Velocity]
	{\includegraphics[width=4.2cm]{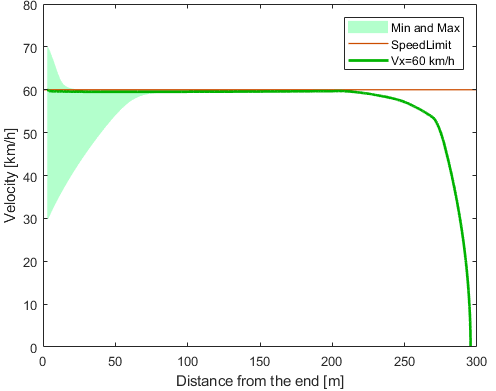}
		\label{5.0a}}
	\subfigure[Distance VS Acceleration]
	{\includegraphics[width=4.1cm]{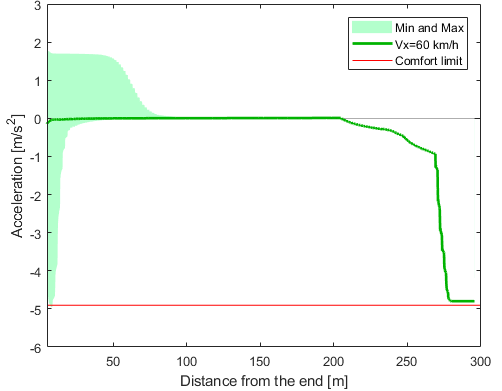}
		\label{5.0b}}
	\caption{Performance of DDPG. The shaded area, dark green and red curve denotes the states visited by agent, reference trajectory and critical value for the penalization.}
	\label{fig5.0}
\end{figure}

Fig. \ref{fig5.0} indicates the performance of the agent trained with DDPG algorithm by applying same conditions as REINFORCE. A completely different driving style is presented not only from the velocity but also from the acceleration figure. The agent is much more sensitive than the one with REINFORCE w.r.t. the speed limit. Especially, it could maintain the velocity slightly lower than the speed limit of the road perfectly (Fig. \ref{5.0a}). This is the performance even cannot be achieved by the expert during the HITL because of imperfectness of human-being. Fig. \ref{5.0b} indicates although the agent is an "aggressive" driver, still he was driving without exceeding the acceleration limit. This is reasonable since the agent doesn't have any pre-knowledge (initial policy parameter) about driving and no one did tell him how to drive beyond the critical value. To summurise, the agent trained by DDPG algorithm successfully achieved all of the goals with much lower computational cost than the REINFORCE.

\section{Conclusion}
In this paper, we presented how to let the vehicle learn the basic behaviors,forward driving and stopping under the traffic rules, via apprenticeship learning and deep reinforcement learning.In particular,we employed GRIL algorithm to recover the reward function and implemented DDPG algorithm to train the agent. Moreover, in order to highlight the performance of DDPG,we employed REINFORCE algorithm in RL step as well.The experimental result shows that our approach successfully trained the agent to drive and stop autonomously by keeping traffic rules and the performance is even better than the expert in the aspect of keeping speed limit.

However, the learnt driving behavior in this study is limited in longitudinal domain. We will introduce steering action and involve other road users to enrich the scenarios in future works.
\bibliographystyle{unsrt}
\bibliography{ictai}

\end{document}